\documentclass{article}
\usepackage{spconf,amsmath,graphicx}
\usepackage[hidelinks]{hyperref}
\usepackage{cite}
\usepackage{amssymb,amsfonts}
\usepackage{textcomp}
\usepackage{booktabs}
\usepackage{multirow}
\usepackage{url}

\newcommand{\pp}{\,\mathrm{pp}}

\title{
GHOST-Q: Towards Studying Grounding Hallucinations Overlooked Under Same-score TradeOffs in Quantized VLMS
}

\name{
Saim Rehman, Muhammad Shafique
}

\address{
\small eBRAIN Lab, Division of Engineering, New York University Abu Dhabi (NYUAD), Abu Dhabi, UAE\\
\small \{sr7849, muhammad.shafique\}@nyu.edu
}

\begin{document}

\maketitle

\begin{abstract}
Post-training quantization of vision--language models (VLMs) is
typically assessed through aggregate task accuracy and memory savings,
but preserving a headline score does not guarantee preservation of
visual grounding behavior. We present GHOST-Q, a cross-precision controlled
 evaluation of three 8B VLM families under
FP16, INT8, and NF4 across utility and hallucination-sensitive benchmarks. Rather than comparing only aggregate accuracy, we pair
FP16 and quantized predictions item-by-item to quantify how compression redistributes grounding successes and failures. Five of six quantized
variants preserve MMStar accuracy within $\pm2$ percentage points, yet 10 of 36 paired effects remain significant after false-discovery-rate
correction, nine on hallucination-sensitive conditions. Same-device A100 profiling further demonstrates that substantial memory reduction does not necessarily mean lower inference latency. Finally, an open-ended AMBER audit reveals strong generation-budget censoring whose
severity varies by architecture and precision. These results show that quantized VLMs should be evaluated jointly for aggregate utility,
grounding reliability, generation behavior, and realized deployment
efficiency.
\end{abstract}
\section{Introduction}

Large vision-language models (VLMs) increasingly serve as general multimodal interfaces, but
their memory footprint motivates aggressive compression. Quantization methods such as LLM.int8(), GPTQ, SmoothQuant, AWQ, and NF4-based QLoRA reduce transformer memory while often retaining conventional task performance \cite{llmint8,gptq,smoothquant,awq,qlora}. In parallel, a distinct literature shows that strong VLMs can hallucinate objects, attributes, relations, or answers that are weakly grounded in visual evidence \cite{pope,amber,hallusionbench,hallucinationsurvey}.
This motivates the following question: \emph{if quantization preserves multimodal utility, does it also preserve the model's pattern of grounding successes and failures?} Moreover, nominally lower precision need not yield lower latency unless the software stack and kernels realize that efficiency. We evaluate this intersection using Qwen3-VL-8B \cite{qwen3vl}, InternVL3-8B \cite{internvl3}, and Idefics3-8B \cite{idefics3}, each in FP16, INT8, and NF4.

\textbf{Motivating case study.} Recent VLM quantization methods such as MBQ and VLMQ improve
low-bit deployment by explicitly accounting for modality imbalance,
token importance, and quantization sensitivity
\cite{mbq,vlmq,vlmbestpractices}. However, their primary validation
still centers on retained multimodal benchmark performance and
compression efficiency, while hallucination benchmarks such as POPE
and AMBER evaluate grounding failures without studying how compression
redistributes them \cite{pope,amber}. \textit{This leaves an important gap:
a quantized model may preserve its headline utility score while changing
which individual examples it grounds correctly. }In our motivating case,
Idefics3-NF4 changes MMStar accuracy by only $+0.40$ pp, yet POPE
accuracy falls by $1.24$ pp, with 257 FP16-correct predictions becoming
wrong versus 145 FP16 errors being corrected
(BH-adjusted $q=3.0\times10^{-7}$). This demonstrates why aggregate
accuracy alone is insufficient to certify grounding reliability after
quantization.

\textbf{Our main contributions are as follows:}
\begin{itemize}
    \item We introduce a controlled cross precision evaluation protocol that compares FP16, INT8, and NF4 VLMs under matched examples, preprocessing, prompts, and decoding conditions across three architectures and multiple utility and grounding-sensitive benchmarks.
    \item  We move beyond aggregate accuracy by introducing an
    item-level paired reliability analysis that distinguishes harmful and
    beneficial precision-induced prediction flips using bootstrap
    confidence intervals, exact McNemar tests, FDR correction, and
    matched-pair effect sizes.
    \item  We jointly characterize behavioral reliability and
    realized deployment efficiency on the same A100 platform, showing when
    memory compression is decoupled from actual inference latency.
    \item We study generation-budget sensitivity in open-ended VLM
    evaluation, showing that fixed decoding budgets can introduce
    architecture- and precision-dependent censoring and therefore confound
    naive hallucination comparisons.
\end{itemize}

\section{Background}

\begin{figure*}[t]
    \centering
    \includegraphics[width=\linewidth]{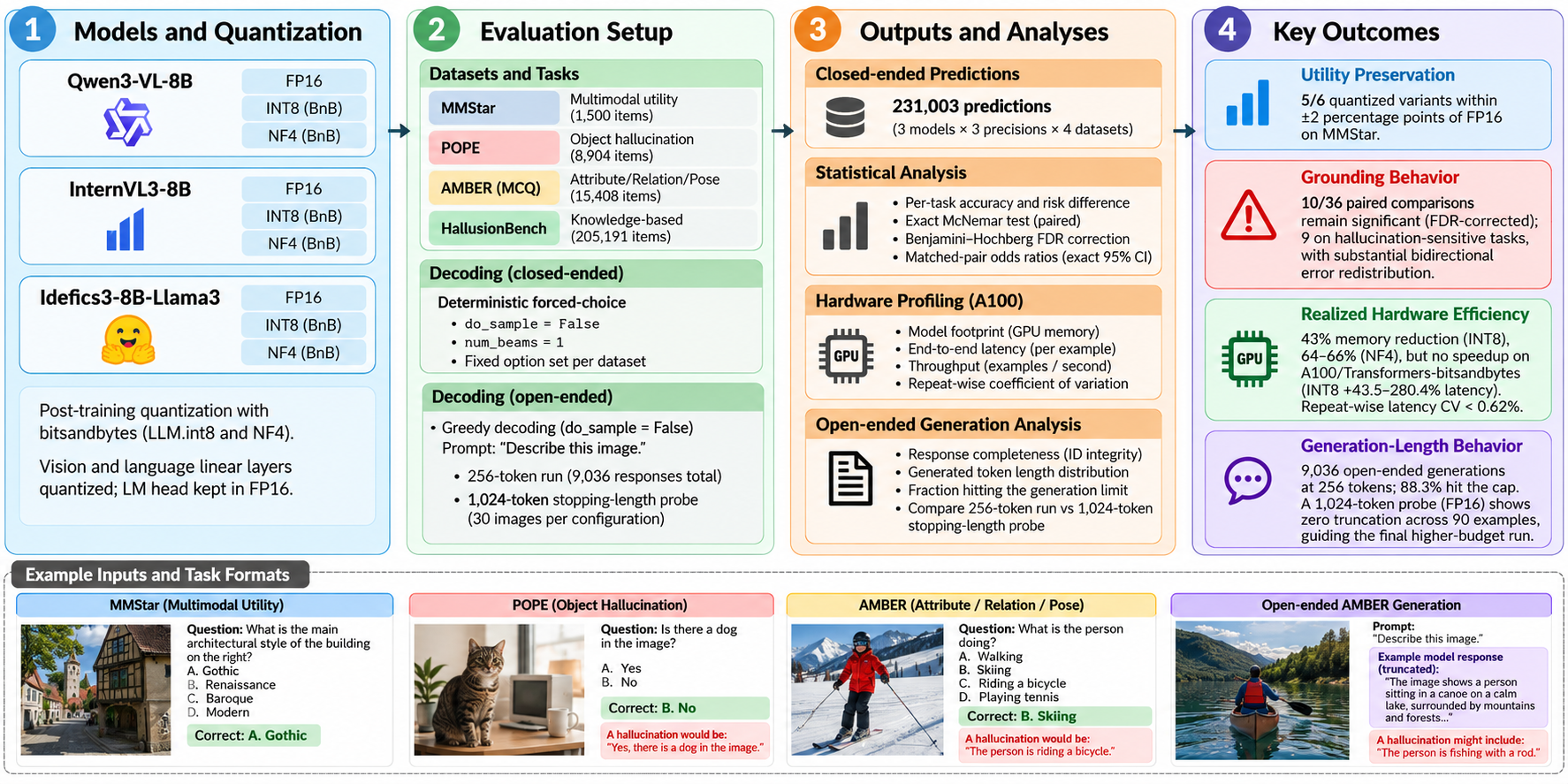}
    \caption{Overview of the Ghost-Q evaluation pipeline and main analyses.}
    \label{fig:methodology}
\end{figure*}

\textbf{Quantization.} To begin, LLM.int8() introduced mixed-precision 8-bit matrix multiplication for large transformers \cite{llmint8}; GPTQ uses approximate second-order information for one-shot low-bit weight quantization \cite{gptq}; SmoothQuant targets W8A8 PTQ \cite{smoothquant}; and AWQ protects activation-salient weight channels \cite{awq}. QLoRA introduced NormalFloat-4 (NF4) and double quantization \cite{qlora}. For multimodal models, Q-VLM targets VLM PTQ through cross-layer dependency modeling \cite{qvlm}; MBQ accounts for different vision/language token sensitivities \cite{mbq}; VLMQ incorporates token importance into a Hessian-based objective \cite{vlmq}; and Das \emph{et al.} systematically study bit width, quantization method, and component sensitivity across multimodal tasks \cite{vlmbestpractices}.

\textbf{Grounding and hallucination.} POPE operationalizes object hallucination through polling-style questions \cite{pope}. AMBER provides LLM-free discriminative evaluation of existence, attribute, and relation hallucination \cite{amber}. HallusionBench probes image-context reasoning under entangled language hallucination and visual illusions \cite{hallusionbench}. Surveys document the broader VLM hallucination landscape \cite{hallucinationsurvey}. Our focus, rather, is complementary: we test whether compression itself changes grounding-sensitive decisions even when aggregate utility is nearly preserved.

\section{Methodology}

\subsection{Controlled Cross-Precision Evaluation}

Ghost-Q compares each quantized VLM against its own FP16 reference
under matched inputs and evaluation conditions. For architecture $m$,
precision $p$, dataset $d$, and example $i$, candidate responses are
scored deterministically from model logits,
\[
\hat y_i^{m,p}
=
\arg\max_{c\in\mathcal C_d}s_{m,p}(c\mid x_i).
\]
We compute accuracy $A_{m,p,d}$ and define the paired precision effect
as
\[
\Delta_{m,p,d}
=
A_{m,p,d}-A_{m,\mathrm{FP16},d}.
\]
All FP16--quantized comparisons use the same examples, preprocessing,
and candidate sets, isolating precision as the experimental factor.
For MMStar, $|\Delta|\leq2$ pp is used only as a practical reporting
tolerance and not as a statistical equivalence test.

\subsection{Paired Grounding-Shift Analysis}

Aggregate accuracy can remain stable even when different examples
change correctness after quantization. We therefore compare FP16 and
quantized predictions item-by-item and count harmful
C$\rightarrow$W and beneficial W$\rightarrow$C transitions. We form
10,000 paired bootstrap resamples for confidence intervals and apply
the exact McNemar test \cite{mcnemar1947} to discordant pairs.
Benjamini--Hochberg correction controls false discovery across all 36
FP16--quantized comparisons \cite{benjamini1995}. Effect direction is
summarized by the matched-pair odds ratio
\[
\mathrm{OR}
=
n_{\mathrm{W\rightarrow C}}/
n_{\mathrm{C\rightarrow W}},
\]
where OR $>1$ indicates more beneficial than harmful flips and
OR $<1$ the reverse.

\subsection{Open-Ended Generation Audit}

Forced-choice evaluation isolates precision-induced decision changes
from free-form decoding. To examine whether the decoding itself
is also precision-sensitive, we additionally evaluate open-ended AMBER
generation under identical prompts, greedy decoding, and a shared
token budget. We record whether each response naturally terminates or
reaches the generation ceiling and use a higher-budget FP16
stopping-length probe to distinguish natural stopping behavior from
budget-induced truncation.

\section{Experimental Setup}
\subsection{Models and Quantization}
We evaluate Qwen3-VL-8B \cite{qwen3vl}, InternVL3-8B
\cite{internvl3}, and Idefics3-8B \cite{idefics3} in FP16,
bitsandbytes LLM.int8(), and NF4. NF4 uses FP16 compute and double
quantization \cite{qlora}, while INT8 uses the mixed-precision
LLM.int8() path \cite{llmint8}. Neither quantized loader requires a
calibration dataset. Module audits show that 92.35--93.38\% of logical
linear weights are quantized; detected vision and language linear
layers are converted while the final LM head remains FP16.

\subsection{Benchmarks}

MMStar provides 1,500 multimodal utility questions
\cite{mmstar}. Grounding-sensitive evaluation uses POPE
(9,000 examples), AMBER existence (4,924), attribute (7,628),
and relation (1,664) items \cite{pope,amber}, and 951
HallusionBench image examples \cite{hallusionbench}. Each
model--precision configuration therefore produces 25,667
forced-choice predictions, giving 231,003 predictions overall.
For open-ended evaluation, we use the 1,004-image AMBER generative
subset with the official ``Describe this image.'' prompt, greedy
decoding, and a common 256-token ceiling. Moreover, Qwen3-VL uses the same fixed 1-megapixel longest-edge image budget
across all three precisions.

\subsection{Hardware Profiling}

All nine configurations are profiled sequentially on the same NVIDIA
A100-SXM4-80GB GPU. After 10 warm-up examples, we perform three
measured repeats over the same 200 POPE inputs, yielding 600 measured
inferences per configuration. We report runtime model footprint, peak
allocated memory, median/p95 end-to-end latency, throughput, and
repeat-wise variability. Power telemetry is treated only as run-level
characterization rather than per-inference energy.

\section{Results}

\subsection{Aggregate Utility is Mostly Preserved}
Table~\ref{tab:main} summarizes utility and grounding-sensitive accuracy. Five of six quantized variants remain within the
$\pm2\pp$ MMStar tolerance; InternVL3-NF4 is the only exception, at $-2.07\pp$ relative to FP16.

\begin{table}[t]
\caption{Accuracy (\%) across utility and grounding-sensitive tasks. ``Macro'' is the unweighted mean of POPE, three AMBER subsets, and HallusionBench.}
\label{tab:main}
\centering
\scriptsize
\setlength{\tabcolsep}{2.15pt}
\begin{tabular}{llrrrrrrr}
\toprule
Model & Prec. & MM* & POPE & A-E & A-A & A-R & Hall. & Macro\\
\midrule
\multirow{3}{*}{Qwen3} & FP16 & 63.53 & 88.68 & 92.73 & 86.88 & 85.22 & 72.77 & 85.25\\
 & INT8 & 64.07 & 88.51 & 93.70 & 86.72 & 85.58 & 73.92 & 85.69\\
 & NF4  & 62.80 & 87.89 & 94.01 & 86.43 & 85.40 & 71.71 & 85.09\\
\midrule
\multirow{3}{*}{InternVL3} & FP16 & 66.47 & 90.87 & 91.96 & 87.02 & 83.35 & 65.93 & 83.83\\
 & INT8 & 65.67 & 91.00 & 92.18 & 86.93 & 83.35 & 64.98 & 83.69\\
 & NF4  & 64.40 & 90.92 & 92.73 & 86.63 & 84.92 & 64.56 & 83.95\\
\midrule
\multirow{3}{*}{Idefics3} & FP16 & 47.33 & 87.43 & 89.40 & 77.31 & 87.02 & 54.15 & 79.06\\
 & INT8 & 46.93 & 86.98 & 89.32 & 76.68 & 87.38 & 54.47 & 78.96\\
 & NF4  & 47.73 & 86.19 & 90.62 & 77.81 & 85.64 & 54.05 & 78.86\\
\bottomrule
\end{tabular}
\end{table}

\subsection{Preserved Averages Hide Paired Grounding Shifts}
\begin{figure}[t]
\centering
\includegraphics[width=\columnwidth]{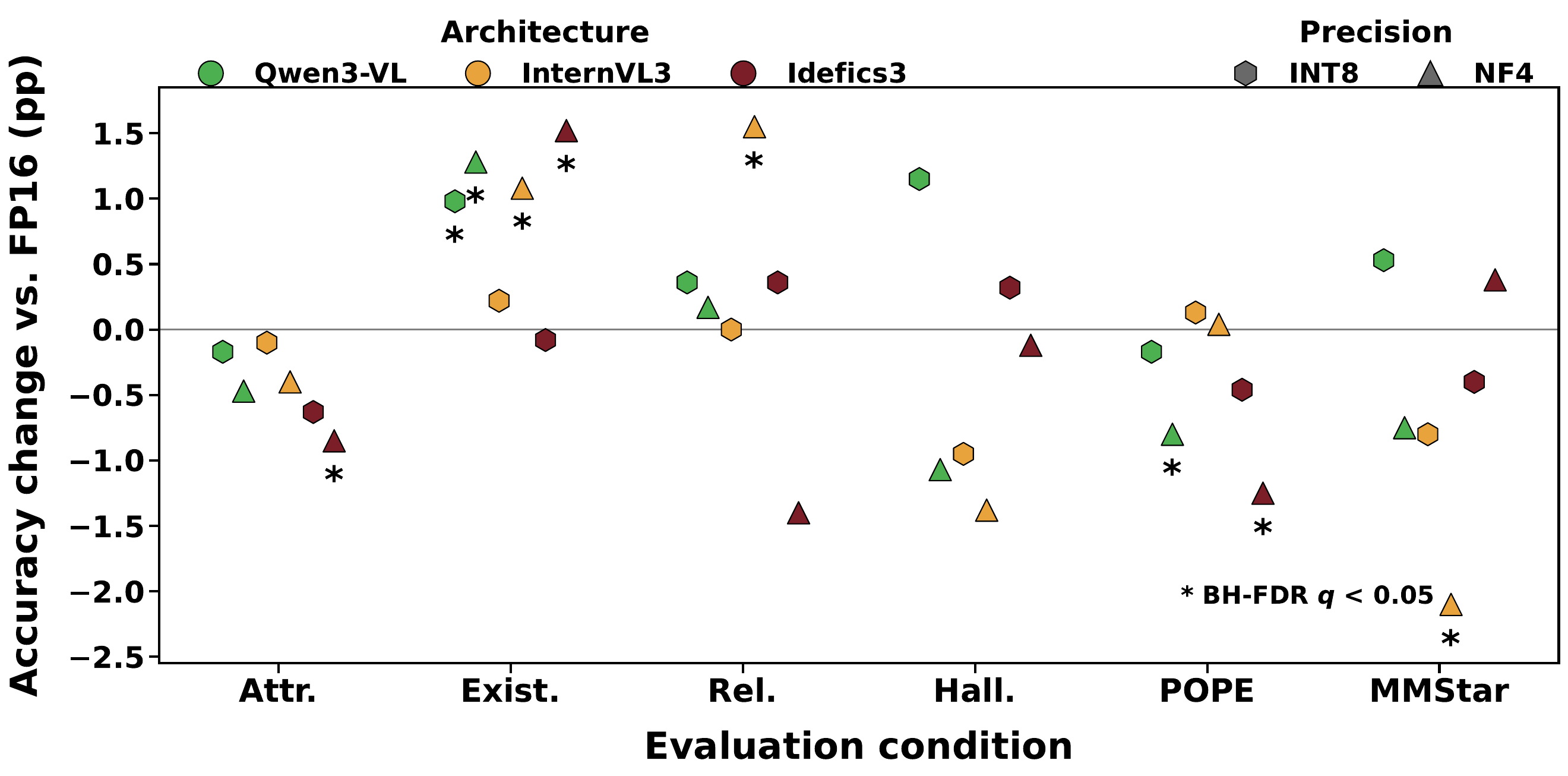}
\caption{Accuracy change relative to each architecture's FP16 baseline. Stars mark comparisons significant after Benjamini--Hochberg correction over all 36 paired tests ($q<0.05$).}
\label{fig:deltas}
\end{figure}

The grounding macro changes by at most $0.43\pp$ from FP16, yet
paired analysis reveals substantial item-level redistribution
(Fig.~\ref{fig:deltas}). Ten of 36 comparisons remain significant
after FDR correction, nine on hallucination-sensitive conditions,
and the effects are bidirectional. Table~\ref{tab:effects} reports
all significant effects; matched ORs range from 13.0 for Qwen3 INT8
on AMBER existence to 0.52 for Qwen3 NF4 on POPE.

\begin{table}[t]
\caption{FDR-significant FP16--quantized paired effects.
$\Delta$ is quantized minus FP16 accuracy (pp).
C$\rightarrow$W/W$\rightarrow$C denote harmful/beneficial flips.
$q$ is the Benjamini--Hochberg-adjusted McNemar $p$-value.
Matched OR $>1$ indicates more beneficial than harmful flips;
OR $<1$ indicates the reverse.}
\label{tab:effects}
\centering
\setlength{\tabcolsep}{2.2pt}
\renewcommand{\arraystretch}{1.03}
\resizebox{\columnwidth}{!}{%
\begin{tabular}{lllrcccr}
\toprule
Model & Task & Prec. & $\Delta$ & C$\rightarrow$W & W$\rightarrow$C &
$q  $ & OR [95\% CI] \\
\midrule
Qwen3     & A-exist & INT8 & +0.97 &   4 &  52 & $<10^{-8}$          & 13.00 [4.78, 49.50] \\
Qwen3     & A-exist & NF4  & +1.28 &  14 &  77 & $<10^{-8}$          & 5.50 [3.09, 10.53] \\
Idefics3  & POPE    & NF4  & -1.24 & 257 & 145 & $3.0\times10^{-7}$  & 0.56 [0.46, 0.69] \\
Qwen3     & POPE    & NF4  & -0.79 & 149 &  78 & $2.6\times10^{-5}$  & 0.52 [0.39, 0.69] \\
Idefics3  & A-exist & NF4  & +1.22 &  67 & 127 & $1.4\times10^{-4}$  & 1.90 [1.40, 2.59] \\
InternVL3 & A-exist & NF4  & +0.77 &  27 &  65 & $5.6\times10^{-4}$  & 2.41 [1.52, 3.92] \\
Idefics3  & POPE    & INT8 & -0.46 &  87 &  46 & $2.5\times10^{-3}$  & 0.53 [0.36, 0.76] \\
InternVL3 & A-rel.  & NF4  & +1.56 &  20 &  46 & $8.4\times10^{-3}$  & 2.30 [1.33, 4.10] \\
Idefics3  & A-attr. & INT8 & -0.63 & 148 & 100 & $1.1\times10^{-2}$  & 0.68 [0.52, 0.88] \\
InternVL3 & MMStar  & NF4  & -2.07 &  74 &  43 & $1.9\times10^{-2}$  & 0.58 [0.39, 0.86] \\
\bottomrule
\end{tabular}%
}
\end{table}

\subsection{Memory Compression Does Not Imply A100 Speedup}
Table~\ref{tab:eff} shows that INT8 reduces model footprint by
42.9--43.8\% and NF4 by 64.3--65.7\%. However, neither improves
median end-to-end latency over FP16 on the tested
A100/Transformers-bitsandbytes stack: INT8 incurs 43.5--280.4\%
higher latency and NF4 +2.2--69.6\%. Repeat-wise median-latency CV
remains below 0.62\%, and NF4 dominates INT8 in both footprint and
median latency for all three architectures. These results characterize
the tested stack rather than quantization algorithms in general.

\begin{table}[t]
\caption{Same-device A100 profile. E2E values are median/p95 latency over 600 measured inferences per configuration.}
\label{tab:eff}
\centering
\scriptsize
\setlength{\tabcolsep}{2.7pt}
\begin{tabular}{llrrrrr}
\toprule
Model & Prec. & Foot. & Peak & E2E$_{50}$ & E2E$_{95}$ & Thr.\\
 & & (GiB) & (GiB) & (ms) & (ms) & ex/s\\
\midrule
\multirow{3}{*}{Qwen3} & FP16 & 16.33 & 16.53 & 76.6 & 86.9 & 13.00\\
 & INT8 & 9.33 & 9.63 & 291.3 & 309.0 & 3.40\\
 & NF4  & 5.83 & 6.18 & 129.8 & 134.7 & 7.70\\
\midrule
\multirow{3}{*}{InternVL3} & FP16 & 14.80 & 16.01 & 354.5 & 363.8 & 3.68\\
 & INT8 & 8.41 & 9.63 & 566.0 & 578.7 & 2.12\\
 & NF4  & 5.22 & 6.55 & 373.9 & 381.1 & 3.43\\
\midrule
\multirow{3}{*}{Idefics3} & FP16 & 15.76 & 16.96 & 416.3 & 537.2 & 2.34\\
 & INT8 & 8.86 & 10.09 & 597.5 & 733.3 & 1.64\\
 & NF4  & 5.41 & 6.73 & 425.5 & 539.2 & 2.30\\
\bottomrule
\end{tabular}
\end{table}

\begin{figure}[t]
\centering
\includegraphics[width=\columnwidth]{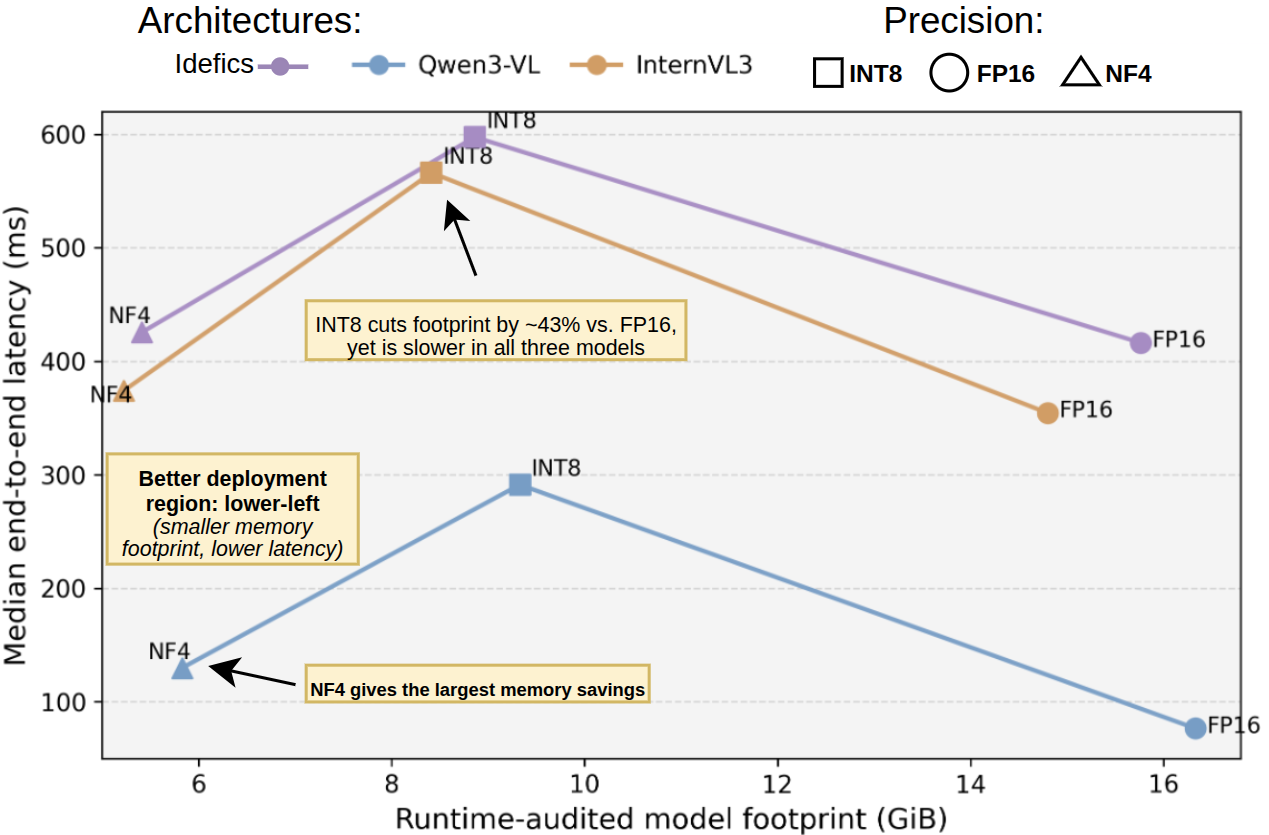}
\caption{Measured footprint--latency trade-off on the same A100 GPU. Quantization consistently moves left (smaller footprint) but not down (faster inference); NF4 dominates INT8 in this two-dimensional efficiency plane for all three architectures.}
\label{fig:memlat}
\end{figure}

\subsection{Open-ended Generation Reveals Budget Sensitivity}

\begin{table}[t]
\caption{AMBER open-ended generation under a common 256-token budget. Entries are responses that hit the token ceiling out of 1,004.}
\label{tab:ambergen}
\centering
\scriptsize
\setlength{\tabcolsep}{4.5pt}
\begin{tabular}{lrrr}
\toprule
Model & FP16 & INT8 & NF4\\
\midrule
Qwen3-VL & 794 (79.1\%) & 768 (76.5\%) & 734 (73.1\%)\\
InternVL3 & 920 (91.6\%) & 913 (90.9\%) & 844 (84.1\%)\\
Idefics3 & 1001 (99.7\%) & 1001 (99.7\%) & 1002 (99.8\%)\\
\bottomrule
\end{tabular}
\end{table}

All nine configurations complete the 1,004-image subset, yielding
9,036 valid responses, but 7,977 (88.28\%) reach the common
256-token ceiling. Censoring varies across both architectures and
precision settings (Table~\ref{tab:ambergen}), showing that
quantization can alter open-ended generation-length behavior under
otherwise identical decoding conditions.

A separate 1,024-token FP16 diagnostic confirms that 256 tokens is
below natural stopping length: none of 90 generations reaches the
1,024-token ceiling, with maximum lengths of 400, 635, and 769 tokens
for Qwen3-VL, Idefics3, and InternVL3, respectively. We therefore
treat the 256-token experiment as a controlled budget-sensitivity
audit rather than an uncensored generative-hallucination estimate.

\section{Discussion and Conclusion}

Across three 8B VLM families, GHOST-Q shows that preserved aggregate
utility does not guarantee preserved grounding behavior or realized
deployment efficiency: quantization can redistribute item-level errors,
and substantial memory savings need not reduce latency on the tested
stack. Open-ended evaluation further reveals precision-dependent
generation-budget censoring. Although limited to three model families,
bitsandbytes weight quantization, and batch-one A100 inference, these
results motivate evaluating quantized VLMs jointly in terms of utility,
paired grounding reliability, generation conditions, and realized
deployment behavior.

\section*{Acknowledgment}
 This work was supported in part by the NYUAD Center for CyberSecurity (CCS), funded by Tamkeen under the NYUAD Research Institute grant G1104. This research was carried out on the High Performance Computing resources at New York University Abu Dhabi.
\vspace{-6pt}
\section*{Generative AI Use Disclosure}

During the preparation of this work, the authors used Generative AI tools (specifically ChatGPT and Grammarly) for language editing, text refinement, and visual refinement of the methodology figure. The authors reviewed and edited all generated or refined content as needed and take full responsibility for the publication’s content.

\bibliographystyle{IEEEtran}
\bibliography{refs}

\end{document}